# ПРАКТИЧЕСКАЯ РЕАЛИЗАЦИЯ САМОБАЛАНИРУЮЩЕГО РОБОТА С ИСПОЛЬЗОВАНИЕМ ИНФРАКРАСНЫХ СЕНСОРОВ

Б. Аубакир, Ж. Каппасов, А. Саудабаев b.aubakir@nu.edu.kz,

zh.kappassov@nu.edu.kz, asaudabayev@nu.edu.kz

*Назарбаев Университет*
Руководитель проекта: ассистент-профессор А. Шинтемиров

*Abstract - The idea of a two wheel self-balancing robot has become very popular among control system researchers worldwide over the last decade. This paper presents a one variant of the implementation of the self-balancing robot using the VEX Robotics Kit.*

**Введение**

В настоящее время самобалансирующие роботы являются очень динамичным направлением развития систем робототехники. Главная целью данных роботов является динамическая поддержка равновесия. Отличие этих роботов заключается в особенности конструкции, в которой отсутствует достаточное количество точек опор для поддержания статической устойчивости. На данный момент, одним из самых известных и узнаваемых примеров систем такого класса является самобалансирующий скутер Segway (Рис. 1), изобретённый Д. Кейменом, работающим в основанной им же компании "FIRST" [1].

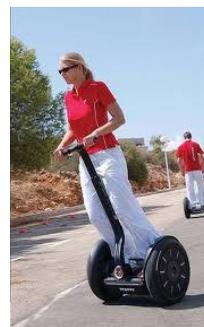

Рис. 1 – мобильная платформа Segway

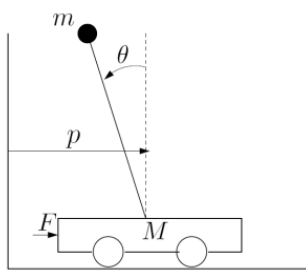

Рис. 2 – Схема перевернутого маятника

Положение равновесия платформы робота наглядно объясняется на примере динамики «перевернутого маятника» из курса физики [2]. На Рис. 2 представлена упрощенная схема самобалансирующей системы. Основание системы в виде тележки массой M балансирует объект массой m, закрепленный на одном конце безмассового стержня, другой конец которого упирается в основание. Таким образом, управляя движением тележки достигается балансирование стержня с объектом в вертикальном направлении.

Переменные $p$ и $\dot{p}$ используются для обозначения текущей позиции и скорости основания соответственно, $\Theta$ и $\Theta'$ обозначают угол наклона и угловую скорость балансируемой части маятника. *F* – сила, приложенная к основанию системы в горизонтальном направлении. Используя вышеперечисленные переменные можно описать динамику системы уравнениями, полученными на основе законов механики Ньютона:

$$(M+m)\ddot{p} + ml\cos\theta\,\ddot{\theta} = -b\dot{x} + ml\sin\theta\,\dot{\theta}^2 + F$$
$$(J+ml^2)\ddot{\theta} + ml\cos\theta\,\ddot{p} = -mgl\sin\theta,$$

где M – масса основания, m и J – масса и момент балансируемого объекта, *l* – расстояние от основания до центра масс объекта, b – коэффициент силы трения между колесами основания и поверхностью, и g – ускорение свободного падения.

Динамические уравнения системы можно переписать в виде уравнений состояния, где *x = (p, Θ, ṗ, Θ')*, входной сигнал *u = F* и результат *y = (p, Θ)*. Уравнения движения после данных преобразований примут вид:

$$\frac{d}{dt}\begin{bmatrix}p\\ \theta\\ \dot{p}\\ \dot{\theta}\end{bmatrix} = \begin{bmatrix}0 & 0 & 1 & 0\\ 0 & 0 & 0 & 1\\ 0 & \frac{m^2gl^2}{J(M+m)+Mml^2} & \frac{-(J+ml^2)b}{J(M+m)+Mml^2} & 0\\ 0 & \frac{mgl(M+m)}{J(M+m)+Mml^2} & \frac{-mlb}{J(M+m)+Mml^2} & 0\end{bmatrix}\begin{bmatrix}p\\ \theta\\ \dot{p}\\ \dot{\theta}\end{bmatrix} + \begin{bmatrix}0\\ 0\\ \frac{J+ml^2}{J(M+m)+Mml^2}\\ \frac{ml}{J(M+m)+Mml^2}\end{bmatrix}u$$

$$y = \begin{bmatrix}1 & 0 & 0 & 0\\ 0 & 1 & 0 & 0\end{bmatrix}x$$

Данную модель балансирующей системы можно использовать для вычисления ответной реакции для стабилизации системы. Имея доступ ко всем перемнным состояния, можно реализовать управление по обратной связи:

$$U = Kx = k_1 p + k_2 \Theta + k_3 \dot{p} + k_4 \Theta',$$

где K ∈ $R^n$ – вектор-строка с четырьмя входными параметрами, известных как коэффициенты усиления при каждом из возможных состоянии. Вследствие такой обратной связи получим управляющую переменную U, которая зависит от конфигурации, баланса и скорости системы. Когда коэффициенты усиления правильно подобраны, система может сохранять себя в равновесии, и малое возмущение верхней части, будет возвращать робот к исходной позиции.

С целью изучения принципа действия подобной системы и ее дальнейшего применения в учебном процессе, был создан упрощенный вариант самобалансирующей платформы мобильного робота. В данной статье кратко описано конструкторская часть проекта, в частности рассмотрены аспекты сборки механической части и представлен алгоритм управления робота.

**Конструкторская часть**

Для сборки робота использовался конструктор роботов Vex Robotics, предназначенный для учебного процесса и имеющий в составе все необходимые копоненты [3]. Необходимые детали для сборки самобалансирующего робота представлены на Рис. 3.

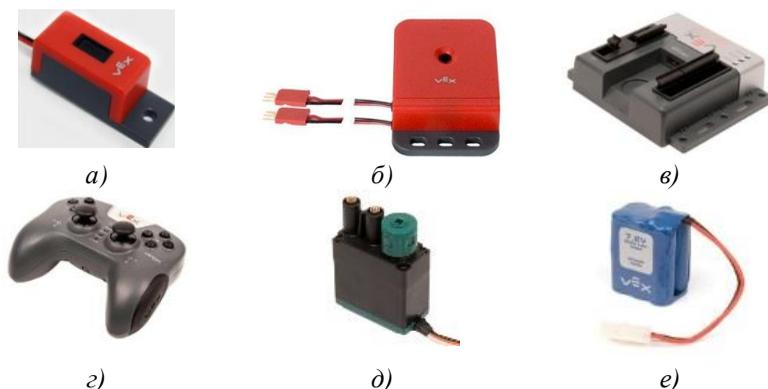

Рис. 3 – инфракрасный сенсор *(а)*, оптический сенсор положения вала *(б)*, микропроцессорный блок управления *(в)*, джойстик управления *(г)*, мотор постоянного тока *(д)*, батарея питания *(е)*.

После сборки металлической рамы корпуса робота, были установлены моторы постоянного тока для привода двух колес по сторонам робота. К осям вращения колес присоединяются оптические сенсоры положения вала, для получения информации об угле поворота и скорости вращения колес. В нижней части рамы устанавливаются инфракрасные сенсоры, для получения

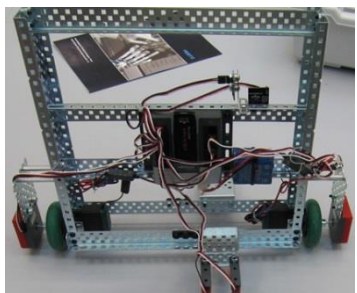

Рис. 4 – фото варианта самобалансирующего робота информации об угле наклона рамы робота. Следующим шагом является установка микропроцессорного блока и подключение всех сенсоров и моторов к нему. Стоит заметить, что устанавливать микропроцессорный блок необходимо ближе к центру, для равномерного распределения веса. Батарея питания, размещается в нижней части рамы робота, с целью смещения центра масс вниз. Инфракрасные сенсоры подключаются к аналоговым входам, в то время как Оптические сенсоры положения колёсного вала подключаются к цифровым входам микропроцессорного блока. Два электромотора также присоединяются через специальный контроллер к процессорному блоку. Последним действием является подключение аккумуляторной батареи к процессорному блоку.

**Управляющая часть**

Схема управления робота (Рис. 5) достаточно проста и включает в себя только четыре сенсора инфракрасного излучения, два оптических сенсора положения вала и пульт, задающий положение робота и представляет собой пропорционально-дифференциальный (ПД) регулятор [4].

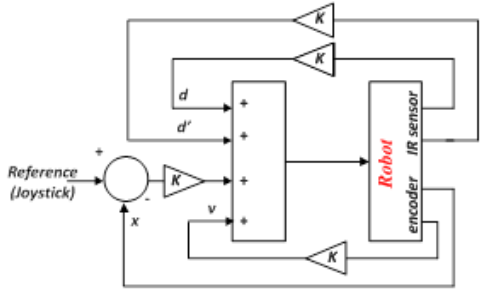

*d – угол наклона робота, d' – скорость изменения наклона, v – угловая скорость вращения колес, x – текущее положение робота относительно первоначального, Reference – установочное положение робота, K – коэффициенты усиления соответствующих переменных.*

Рис. 5 – Структурная схема блока управления робота

Данные с инфракрасных сенсоров информируют о расстоянии сенсора до поверхности (Рис. 6). При предварительном калибровании и путем вычета первоначального смещения, с помощью простого алгоритма сочетания сенсоров возможно получить информацию о наклоне всего робота от положения равновесия:

$$d = d_{front} - d_{back}, \qquad (1)$$

где $d_{front}$ и $d_{back}$ – соответственно расстояние переднего и заднего сенсоров до поверхности, $d$ – отражает угол наклона робота от положения равновесия. Следует отметить, что $d_{front}$ и $d_{back}$ среднеарифметические показания с двух передних и двух задних сенсоров.

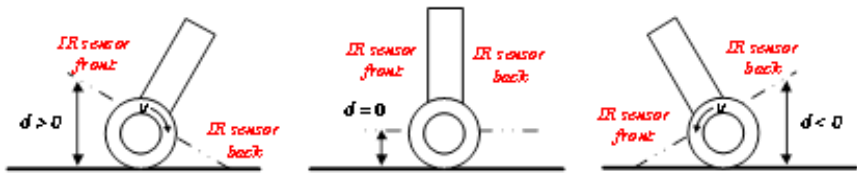

Рис. 6 – Положение робота и данные с инфракрасных сенсоров.

Данные с оптического сенсора положения вала отображают угол поворота вала, к которому прикреплено колесо робота, от первоначального положения. При заданном диаметре колеса данные с валового сенсора представляют собой пройденное расстояние, которое на рисунке 5 *угол наклона робота, d' – скорость изменения наклона, v – угловая скорость вращения колес, x – текущее положение робота относительно первоначального, Reference – установочное положение робота, K – коэффициенты усиления соответствующих переменных.*

Рис. 5 обозначено символом *x*.

Для получения скорости движения робота и скорости изменения угла наклона необходимо произвести операцию дифферинцирования. В микропроцессорной технике это реализуется с помощью простейшего дифференциатора [4]:

$$y = \alpha * y - (1-\alpha) * \frac{(x_i - x_{i-1})}{t_d}, \qquad (2)$$

где $y$ – производная величина, $\alpha$ – константа, равная 0,99, $t_d$ – время задержки, равная 1 мс, $x_i$ – текущее и $x_{i-1}$ – предыдущее значения дифференцируемой величины.

Таким образом, подставляя значения сенсоров положения вала вместо переменной *x* в формуле (2), можно получить скорость движения робота, а при подстановке переменной *d* – скорость изменения угла наклона.

Суть балансирования робота состоит в управлении моторами робота таким образом, что при отклонении робота от положения равновесия, необходимо вращать колеса в направление отклонения (Рис. 6). Для того чтобы робот балансировал в заданной точке поверхности, по цепи обратной связи с отрицательным знаком в регулятор подается значение положения вала в то время как задающее значение положения задается с джойстика с положительным знаком. Ошибка с умножающим коэффициентом *k* подается на сумматор, как и данные об отклонении, скорости отклонения и скорости вращения колес с положительным знаком (Рис. 5).

**Полученные результаты и планы**

Вышеописанный алгоритм управления был реализован в среде разработки RobotC с использованием С-подобного языка программирования для конструктора Vex Robotics. Коэффициенты регулирования подбирались опытным путем в ходе экпериментальных исследований прототипа робота. Работу прототипа робота можно увидеть по следующей ссылке http://youtu.be/iJ1Ab-Jl1a8 .

В целом, робот демонстрирует устойчивость и управляемость. Однако, вследствие наличия значительного люфта моторов и использования инфракрасных сенсоров, наблюдается некоторая неточность и видимые осцилляции вокруг заданного положения. В дальнейшем запланировано улучшение функционирования робота с введением сенсоров гироскопа и акселерометра, что является сравнительно более сложной задачей. Также рассматриваются варианты использования фильтра Кальмана или комплементарного фильтра в системе управления и проектирования более сложных регуляторов (LQR регуляторы) с использованием точной математической модели робота.

**Заключение**

В рамках проекта был создан прототип самобалансируемого робота, который может быть использован в учебных и исследовательских целях. Описаны конструкторская и программная части робота, представлены детали алгоритмов самобалансирования и дистанционного управления роботом с помощью джойстика. Помимо применяемого в данной работе ПД регулятора, разработанный прототип робота позволяет реализовать ПИД, ПИ и другие более сложные регуляторы управления, а также сенсоры гироскопа и акселерометра, что будет применяться для практических занятий в рамках курсов теории систем управления линейными системами, сенсоры и обработка информации и т.д.

В перспективе практический опыт, полученный в рамках реализации данного проекта, будет использован в научно-исследовательской работе Департамента робототехники и мехатроники Назарбаев Университета для разработки и реализации систем управления более сложных робототехнических систем.